\documentclass[conference]{IEEEtran}
\IEEEoverridecommandlockouts
\usepackage{cite}
\usepackage{amsmath,amssymb,amsfonts}
\usepackage{algorithmic}
\usepackage{graphicx}
\usepackage{textcomp}
\usepackage{xcolor}
\usepackage[caption=false]{subfig}
\def\BibTeX{{\rm B\kern-.05ecommmm{\sc i\kern-.025em b}\kern-.08em
    T\kern-.1667em\lower.7ex\hbox{E}\kern-.125emX}}

\makeatletter
\newcommand{\linebreakand}{%
  \end{@IEEEauthorhalign}
  \hfill\mbox{}\par
  \mbox{}\hfill\begin{@IEEEauthorhalign}
}
\makeatother

\begin{document}

\title{ArUcoGlide: a Novel Wearable Robot for Position Tracking and Haptic Feedback to Increase Safety During Human-Robot Interaction
}

\author{\IEEEauthorblockN{Ali Alabbas}
\IEEEauthorblockA{\textit{Skolkovo Institute of Science  
and Technology}\\
Moscow, Russia \\
ali.alabbas@skoltech.ru}% or ORCID}
\and
\IEEEauthorblockN{Miguel Altamirano Cabrera}
\IEEEauthorblockA{\textit{Skolkovo Institute of Science 
and Technology}\\
Moscow, Russia \\
miguel.altamirano@skoltech.ru} %0000-0002-5974-9257}
\linebreakand
\IEEEauthorblockN{Oussama Alyounes}
\IEEEauthorblockA{\textit{Skolkovo Institute of Science  
and Technology}\\
Moscow, Russia \\
oussama.alyounes@skoltech.ru}% or ORCID}
\and

\IEEEauthorblockN{Dzmitry Tsetserukou}
\IEEEauthorblockA{\textit{Skolkovo Institute of Science
and Technology}\\
Moscow, Russia \\
d.tsetserukou@skoltech.ru}% or ORCID}

}

\maketitle

\begin{abstract}
The current capabilities of robotic systems make human collaboration necessary to accomplish complex tasks effectively.  In this work, we are introducing a framework to ensure safety in a human-robot collaborative environment. The system is composed of a wearable 2-DOF robot, a low-cost and easy-to-install tracking system, and a collision avoidance algorithm based on the Artificial Potential Field (APF). The wearable robot is designed to hold a fiducial marker and maintain its visibility to the tracking system, which, in turn, localizes the user's hand with good accuracy and low latency and provides haptic feedback to the user. The system is designed to enhance the performance of collaborative tasks while ensuring user safety. Three experiments were carried out to evaluate the performance of the proposed system. The first one evaluated the accuracy of the tracking system. The second experiment analyzed human-robot behavior during an imminent collision. The third experiment evaluated the system in a collaborative activity in a shared working environment. The results show that the implementation of the introduced system reduces the operation time by 16\% and increases the average distance between the user's hand and the robot by 5 cm.

\end{abstract}


\section{Introduction}
% from Oussama

The number of robots in the industry is increasing globally, reaching almost 3.5 million robots in factories in 2021, according to the International Federation of Robotics (IFR) \cite{IFR}. Robots are considered a helpful tool for carrying out repetitive tasks and operating in hazardous environments, while human dexterity can be harnessed in the operation. However, robots are mostly being implemented in the industry as tools, not as companions for humans. To achieve collaboration between humans and robots, the safety of users has to be guaranteed.

Sharkawy et al. introduced a system that consists of a force sensor located at the robot tool center point (TCP) to detect collisions using neural networks (NN) algorithms \cite{paper3}. However, this system reacts to a collision only after it occurs.

Numerous frameworks regarding collision avoidance algorithms were developed to achieve human-robot collaboration. The simplest strategy is to stop the robot when humans are near the robot's path area \cite{paper1, paper2}. This behavior results in a longer operation time due to the continuous cessation of the movement of the robot. 
Real-time obstacle avoidance algorithms are used to avoid collisions between robots and obstacles when the obstacle's position is known to the system. 

\begin{figure}[t!]
  \centering
  \includegraphics[width=0.46\textwidth]{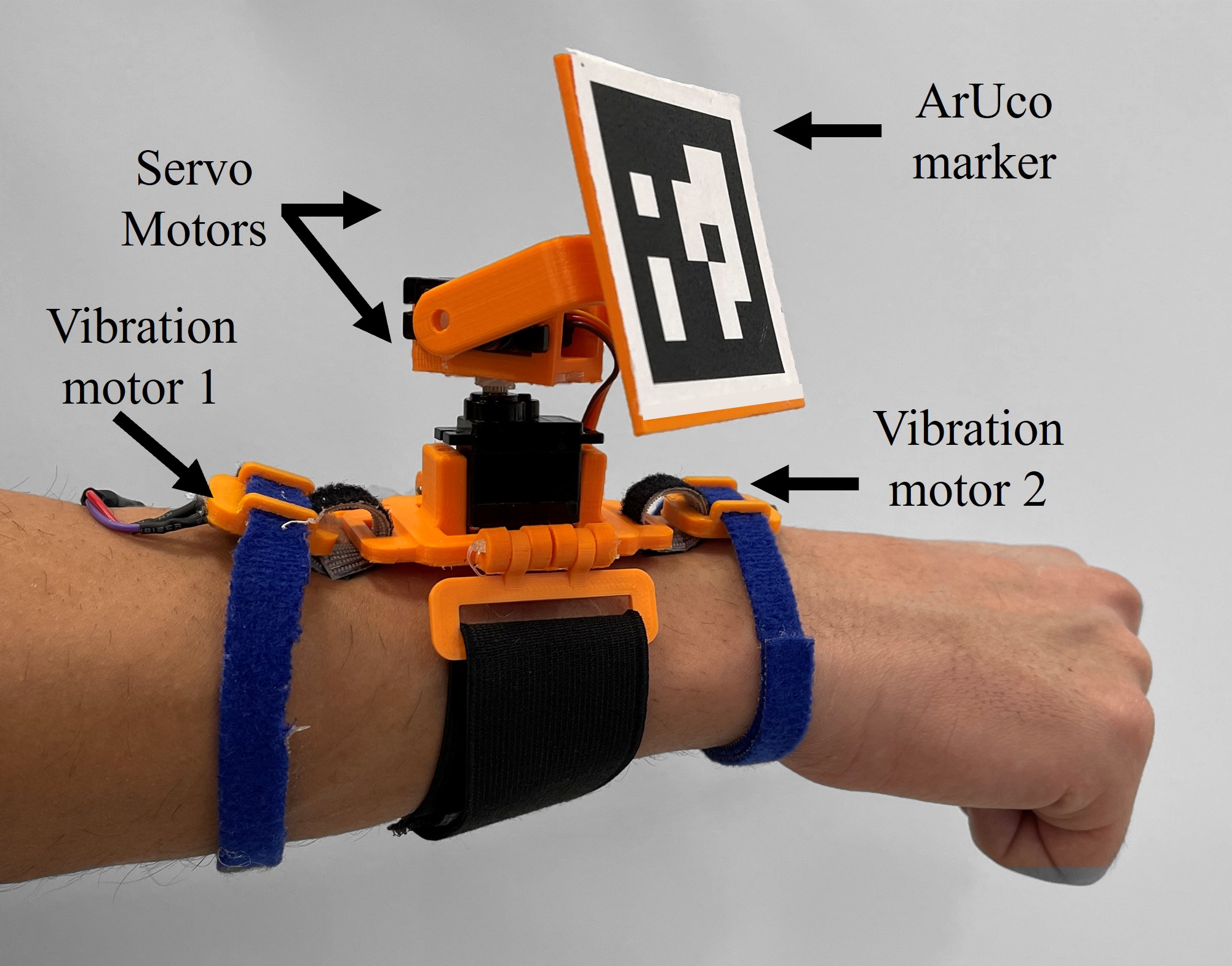}
  \qquad
  %\vspace*{-5mm}
  \caption{ArUcoGlide, a novel wearable robot for position tracking and haptic feedback. An ArUco marker is located at the end effector to adjust its orientation, and the two vibration motors provide haptic feedback to the user. 
  }
  \label{fig:Device_arm}
  \vspace{-3mm}
\end{figure}

Khatib introduced the collision avoidance algorithm based on Artificial Potential Fields (APF), where the target point is considered the center of an attractive field and the obstacles are considered centers of repulsive fields \cite{Khatib}. 

Several studies focused on enhancing the APF algorithm to overcome the local minimal short-come. Pan et al. added an intermediate target point on the path of the robot to reach the final goal \cite{EnhancedAPF}. While some studies developed an algorithm by activating the repulsive APF depending on the direction of the movement of the robot and its velocity relative to the obstacles \cite{cobotgear}, \cite{HR-InteractionSafety}.

Environment perception is an essential step in detecting the position of obstacles in the robotic environment before applying collision avoidance algorithms. Many optical motion capture systems (mocap) are being used, such as VICON \cite{Vicon1}, \cite{Vicon2}, or OptiTrack \cite{OptiTrack}. These types of mocap systems are distinguished by their high accuracy and short time delay. Nevertheless, they are expensive and require powerful computers for data processing. Fiducial markers are commonly used during object manipulation for the identification, detection, and localization of different objects \cite{FiducialMarker2}. Their costly-efficient properties and high accuracy helped to implement them in wide application areas, including industrial systems, augmented reality, robot navigation, human-robot interaction (HRI), and others. The AprilTag, ARTag, STag, ArUco, and CALTag are one of the most commonly used fiducial markers. These markers perform differently depending on their occlusion \cite{Tagscomparison}. STag and ArUco show high precision in position and rotation compared to the other types of markers \cite{FiducialMarker1}.

ArUco markers were first proposed as a method for camera pose estimation \cite{FirstArUcoMarker}. However, they started to be used for different purposes, especially in the HRI field \cite{HR-InteractionSafety}, \cite{HR-ArUco1}. Rajendran et al. developed a framework for HRI integrating ArUco markers for an assembly task \cite{HR-framework}. A liver surgical support system was developed using ArUco markers by Koeda et al. \cite{ArUcoInSurgery}, where a surgical knife was attached to two ArUco markers in order to avoid the collision of the surgical knife with the liver. However, the main problem with using ArUco markers for pose estimation is the occlusion and the decreasing of their detection rate according to their orientation \cite{FiducialMarker1}.

This paper presents a novel system framework to guarantee safety during human-robot collaborative tasks. The framework is composed of a novel wearable robot that holds an ArUco marker for the position estimation of the user's hand, dubbed ArUcoGlide; haptic feedback provided by the wearable robot to inform the users about hazardous events; and a collision avoidance algorithm based on APF. The wearable robot is composed of two servo motors, giving the ArUco marker the freedom to rotate around two different axes with the objective of modifying its position and keeping a specific orientation to the camera, avoiding occlusions, and increasing its detection rate. To provide haptic feedback to users, two vibration motors are embedded into the device. Experiments were conducted to evaluate the effectiveness of the ArUcoGlide wearable robot and the behavior of the full system during a human-robot collaborative task.

\section{System Overview} \label{System Integration}

The proposed system comprises three distinct components: 1) a tracking unit consisting of a computer and a camera that continuously captures a live video stream of the workspace; 2) a wearable robot, dubbed ArUcoGlide, that adjusts its orientation to ensure the visibility of the ArUco marker and to address occlusion by objects within the environment, with the additional functionality of providing haptic feedback in the event of a hazardous situation; and 3) a collision avoidance controller that governs the movement of the robot, which in our case is a 6 DoF collaborative robot UR10, to avoid potential collisions with the user's hand in the workspace.

In the following subsections, we will describe each of the system components separately.

\subsection{ArUcoGlide wearable robot}
\begin{figure}[h]
 \centering%
 \subfloat[Isometric view of the 3D model.]{%
         \label{fig:Isometric}%
         \includegraphics[width=0.7\linewidth]{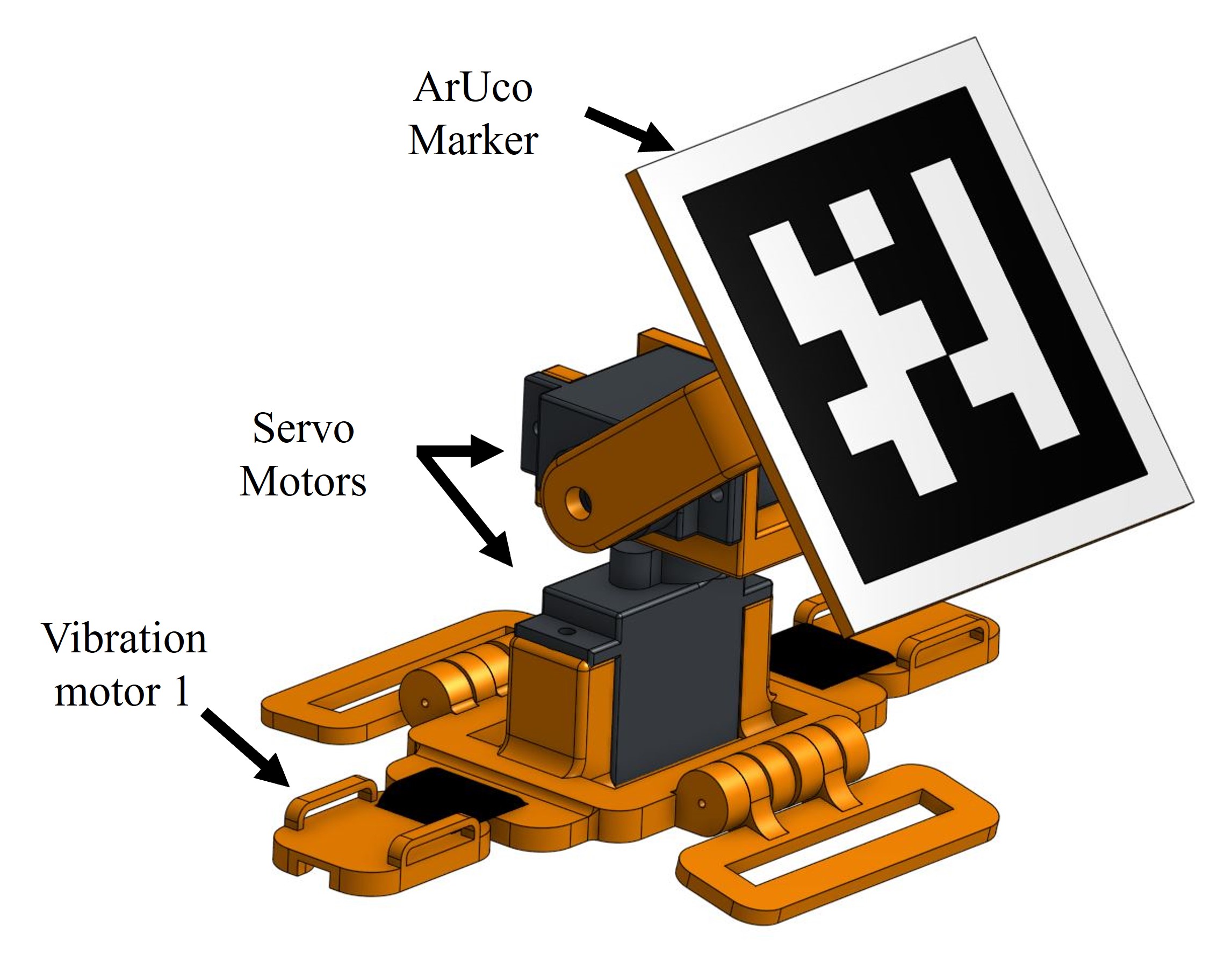}%
} \qquad
  \centering%
 \subfloat[Front view of the 3D model.]{%
         \label{fig:Front}%
         \includegraphics[width=0.6\linewidth]{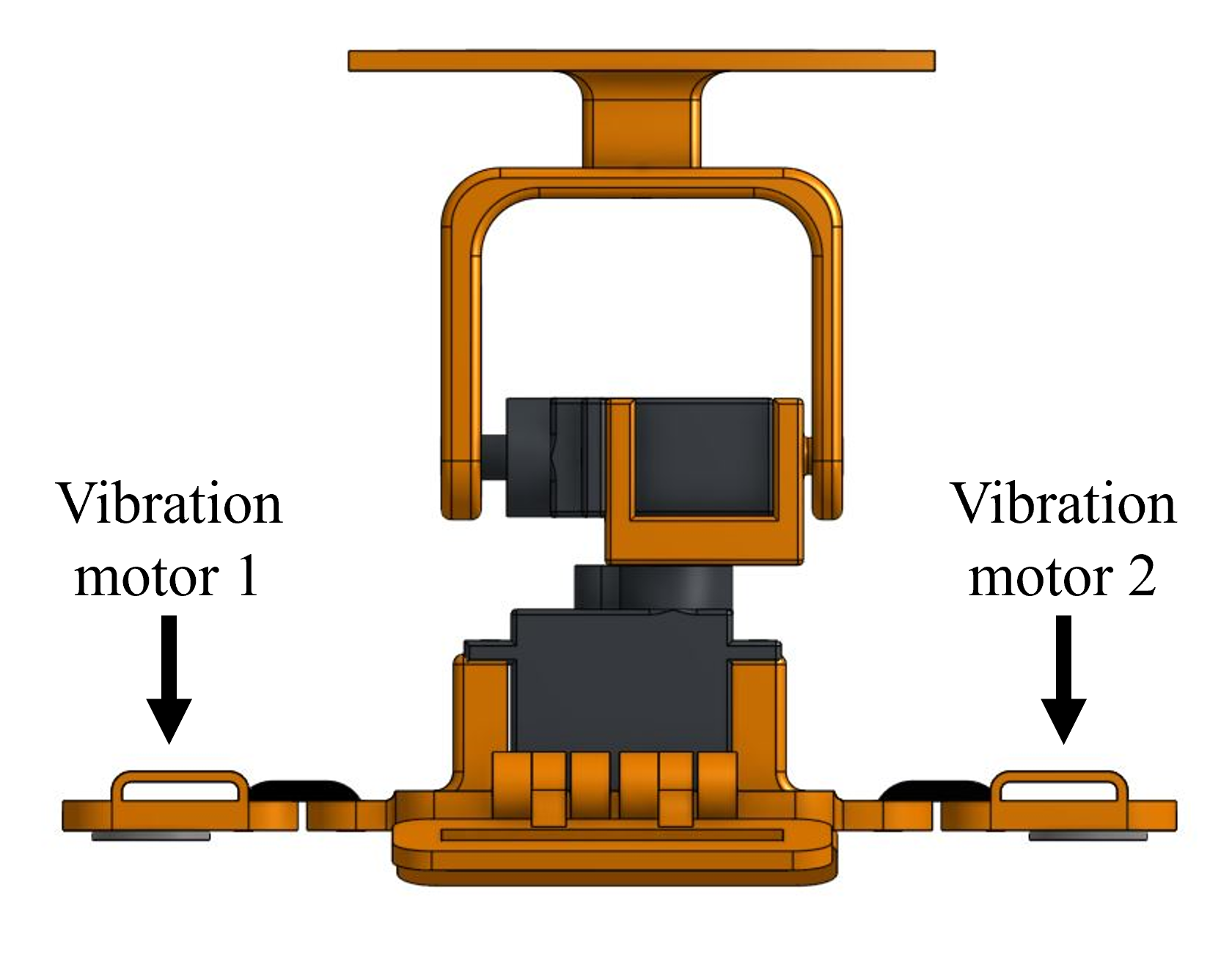}%
} \qquad
  \caption{3D CAD model of the ArUcoGlide wearable robot.}
  \label{fig:GlideQR}
\end{figure}

ArUcoGlide is a novel wearable two-degrees-of-freedom (2-DOF) robot that holds an ArUco marker at its end effector with the objective of estimating the position of the user and ensuring safe HRI. It consists of an ESP32 microcontroller and two SG90 servo motors, as shown in Fig. \ref{fig:Device_arm}. The links and holders were designed and 3D-printed with PLA material; the 3D CAD model is shown in Fig. \ref{fig:GlideQR}. The position of the servo motors and the vibration of the vibration motors are controlled via Bluetooth from a base computer. ArUcoGlide primary function is to maintain the visibility of the ArUco marker to the tracking system by continuously adjusting the motors' angular position to hold the marker in a fixed orientation to the camera frame.

To enhance user safety, ArUcoGlide is equipped with two vibration motors located on the extreme sides of the device. The vibration motors generate stimuli to alert the user in dangerous situations, i.e., when the robot is too close to the user's hand or when the marker is not visible.

ArUcoGlide was designed to fit comfortably on the user's forearm, allowing unrestricted user movement during object manipulation tasks. This design feature ensures that the device does not interfere with the user's natural range of motion and provides a seamless user experience.

\subsection{Tracking system}
When collaborating with a robotic platform, it is essential to make sure the safety of the human operator is never compromised. To achieve this, it is crucial to track the real-time position of the operator within the working space. Our proposed tracking system is both cost-efficient and easy to install, utilizing an ArUco marker to track the operator's hand. The marker's position is continuously sent to the collision avoidance controller, allowing real-time monitoring of the operator's movements within the workspace. The system comprises an HD webcam C930e from Logitech mounted on a stand that can be adjusted to capture different angles of the workspace, providing greater flexibility for the user.
To determine the position of the user relative to the UR10 robot, we first need to know the camera's position relative to the robot. This involves finding the transformation matrix, denoted as $T_B^C$, that represents the relationship between the camera and the UR10 robot's coordinate systems. We place a reference marker at a known location relative to the UR10 base, with a transformation matrix from the marker to the UR10 coordinate system $T_B^{M}$. When this marker is detected from the camera, we can acquire the transformation from the marker to the camera $T_C^{M}$. As a result, we obtain the transformation between the camera and robotic arm coordinate systems $T_B^C$ as follows:

\begin{equation}\label{coordinates_transform}
T_B^C = T_B^{M} (T_C^{M})^{-1}.
\end{equation}
This process is required only once during the initial setup of the system. However, if the camera's position or orientation is altered, the process needs to be repeated to derive the correct transformation matrix. Once the transformation matrix $T_B^C$ is determined, we can utilize it to track the position of the ArUco marker that is attached to the user's hand, enabling us to locate the marker within the UR10 robot's coordinate system.

 \begin{figure}[ht]
  \centering
  \includegraphics[width=0.30\textwidth]{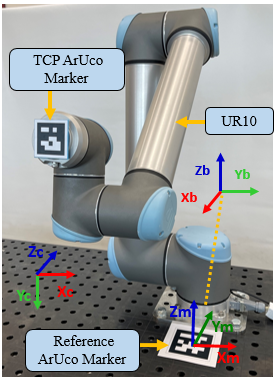}
  \qquad
  \caption{UR10 Robot Setup with Reference Marker for Motion Tracking: Evaluation of Tracking Accuracy using TCP Marker in Experiment 1} 
  \label{fig:tracking_exp}
  \vspace{-3mm}
\end{figure}

\subsection{Control Algorithm}

The control algorithm was proposed by us in \cite{cobotgear} and \cite{9551579}. In this subsection, we will give a brief review of the control algorithm. The role of the controller is to guide the robot's TCP position ${x}_{R}$ to the goal position ${x}_{G}$, avoiding an obstacle at position ${x}_{O}$ based on the APF approach, which employs virtual repulsive and attractive fields associated with obstacles and targets faced during robot movement.

 \begin{figure}[ht]
  \centering
  \includegraphics[width=0.45\textwidth]{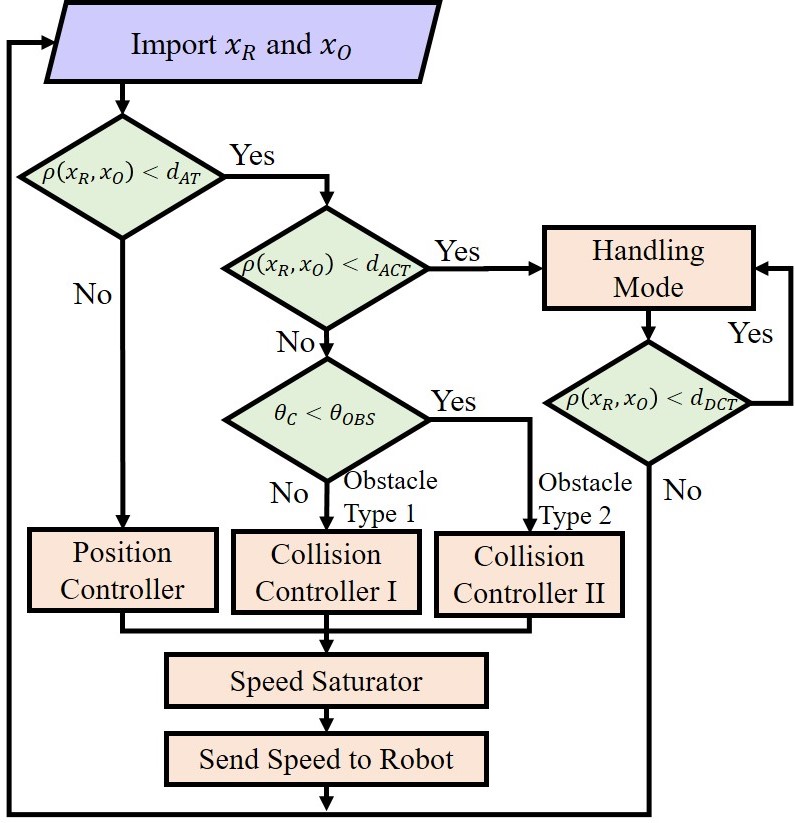}
  \qquad
  \caption{Behavior tree of the control architecture. Position controller and collision controller I and II are defined by (\ref{eq:C}), (\ref{eq:CI}), and (\ref{eq:CII}), respectively. } 
  \label{fig:Three}
\end{figure}

We apply the algorithm represented in the behavior tree in Fig. \ref{fig:Three}. The decisions are made based on the distance between the robot TCP and the obstacle $d_{RO} = \rho(\boldsymbol{x}_{R}, \boldsymbol{x}_{O})$. The behavior tree is divided into three cases. 

\subsubsection{Case 1} The obstacle is far from the robot's TCP position. The robot-obstacle distance $d_{RO}$  is greater than the avoidance threshold distance $d_{AT}$ (the no-avoidance region). Therefore, the risk of a collision is low. The algorithm employs the position controller represented in equation (\ref{eq:C}):
\begin{equation}\label{eq:C} \boldsymbol{\dot{q}}_{PC} = J^{-1} \cdot ( \boldsymbol{\dot{x}}_{G} + k_{PC1} \tanh{(k_{PC2} \cdot \boldsymbol{e})})   , \end{equation}
where $\dot{q}_{PC}$ is the articulation speed of the TCP, $J$ is the Jacobian matrix, $\boldsymbol{e}$ is the position error between the goal position ${x}_{G}$ and the robot TCP position ${x}_{R}$, $k_{PC1}$ and $k_{PC2}$ are the calibration parameters of the collision avoidance controller. 

\subsubsection{Case 2} 
\begin{figure}[ht]
  \centering
  \includegraphics[width=0.45\textwidth]{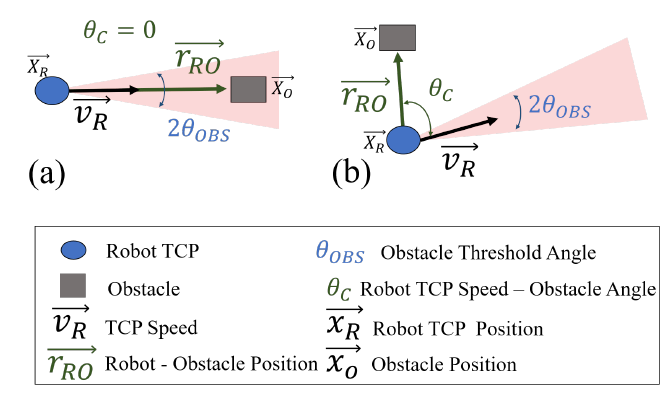}
  \qquad
  \caption{a) obstacle type 1: the angle between the robot-obstacle position vector and the TCP velocity vector is less than the threshold $\theta_{OBS}$. b) obstacle type 2: the angle between the robot-obstacle vector and the TCP velocity vector is greater than $\theta_{OBS}$ } 
  \label{fig:case2}
\end{figure}

The obstacle is inside the avoidance area ($d_{RO} < d_{AT}$), but not in the critical area ($d_{RO} > d_{ACT}$). In such a case, there is the risk of a collision. The algorithm classifies the obstacle as one of two different types according to the angle between the TCP velocity vector and the robot-obstacle vector $\theta_{C}$. When the angle $\theta_{C}$ is less than a threshold value $\theta_{OBS}$ the obstacle is type 1, and the collision is imminent if the robot's movement continues in the same direction. If the angle $\theta_{C}$ is greater than the threshold angle $\theta_{OBS}$, the obstacle is type 2, and this ensures the robot will not collide with the obstacle. Equations (\ref{eq:CI}) and (\ref{eq:CII}) define the obstacle avoidance control in the case of obstacle types 1 and 2, respectively: 
 \begin{equation}
\label{eq:CI}
\dot{q}_{CCI} = J^{-1}( \boldsymbol{v}_{PC} ( 1 - e^{-\tau d_{RO}}) + \boldsymbol{v}_{rep_1}  e^{-\tau d_{RO}})   , 
\end{equation}
where $\tau$ is the space-null attenuation constant, ${v}_{PC}$ and ${v}_{rep_1}$ are the speed of the position controller and speed of the collision controller I, respectively, and ${v}_{rep_1}$ is composed of three repulsive forces to change the direction of the robot TCP and avoid the obstacle \cite{cobotgear}.
\begin{equation}
\label{eq:CII}
\dot{q}_{CCII} = J^{-1}( \boldsymbol{v}_{PC} ( 1 - e^{-\tau d_{RO} }) + \boldsymbol{v}_{rep_2}  e^{-\tau d_{RO}}) ,
\end{equation}
where $\tau$ is the space-null attenuation constant, ${v}_{rep_2}$ is the speed of the collision controller II that is composed of only a normal repulsive force \cite{cobotgear}. Fig.\ref{fig:case2} illustrates the two mentioned obstacle types.

\subsubsection{Case 3}  When $d_{RO}$ is smaller than the critical threshold distance of activation $d_{ACT}$ and the obstacle is exceptionally close to the robot TCP. Therefore, the risk of collision is high. The algorithm initiates a Free Drive Control Mode (FDCM). In the case of FDCM, the robot stops, and the user can modify the robot's position by pulling or pushing the end effector. The system deactivates FDCM when the robot-operator distance $d_{RO}$ is larger than the critical threshold value for deactivation $d_{DCT}$.

In this work, we study the influence of the ArUcoGlide wearable robot during a human-robot interaction task. We assume that adjusting the marker orientation to stay visible to the camera reduces the marker occlusion moments, which will increase operation time. We also analyze the influence of the haptic feedback provided by the proposed device to alarm the user in hazardous situations. It will reduce the robot's collision avoidance path due to informing the user when the robot is nearing a critical distance. We propose four modes of operation for the system:
\begin{itemize}

\item Mode 1: When the distance from the robot TCP to the user's hand $d_{RO}$ is larger than the avoidance threshold distance $d_{AT}$ (the no-avoidance region). In this case, the ArUcoGlide is only adjusting the marker orientation to the camera to stay around the desired orientation.

\item Mode 2: When the distance from the TCP to the user's hand $d_{RO}$ is between the critical distance of activation $d_{ACT}$ and the avoidance threshold distance $d_{AT}$ (the avoidance region). In this case, one vibration motor will be activated to inform the user of the collision avoidance path activation. The device is also adjusting the marker orientation at the same time.

\item Mode 3: When the distance from the TCP to the user's hand $d_{RO}$ is smaller than the critical distance of activation $d_{ACT}$ (the critical region). In this case, the two vibration motors will be activated to inform the user of an imminent collision. The robot is stopped, and the FDCM is activated. 

\item Mode 4: When the marker is occluded or is not visible. In this case, the robot is stopped, and the FDCM is activated. The two vibration motors will be activated to inform the user that the robot is stopped and that his hand is not visible. 

\end{itemize}

In the next section, we will start evaluating the overall performance of the proposed system.

\section{Experimental Evaluation} \label{ALGORITHM CALIBRATION EXPERIMENTS}

Two experiments were carried out to evaluate the performance of the system. The first experiment evaluated the accuracy of the tracking system and its ability to track a moving target. The second experiment analyzes the robot's behavior during an imminent collision. 

\subsection{Experiment 1}
The present experiment aims to evaluate the precision of the detection system in order to determine a safe zone for the moving operator's hand. To accomplish this goal, we rely on the UR10 universal robot arm, which can accurately localize its TCP and maintain it at a fixed position under nominal payload conditions, as specified by the manufacturer.
To assess the detection system's accuracy, we fixed an ArUco marker to the TCP of the UR10 robot as shown in Fig. \ref{fig:tracking_exp}, which is detectable by the camera. At each time step, we extract the TCP coordinates using the tracking system and compare them with the coordinates calculated through the UR10 forward kinematics. This enables us to compute the position error for each axis and determine the tracking system's precision in detecting the marker.
We programmed the UR10 robot arm to sweep through different positions at a velocity equal to $0.1 \ m/s$ while tracking the marker affixed to the moving TCP. %We recorded both the TCP coordinates provided by the tracking system and those obtained from the robotic arm forward kinematics.

\begin{figure}[h!]
  \centering
   \includegraphics[width=0.5\textwidth]{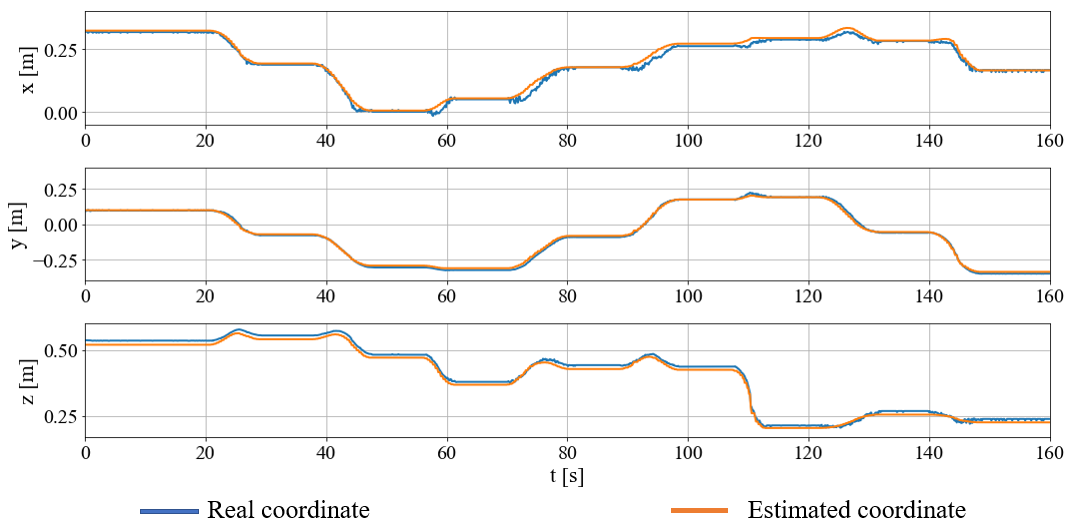}
  \qquad
  \vspace*{-5mm}
  \caption{Experiment 1: Position error on each axis when the TCP is sweeping between different positions to cover the camera frame.}
  \label{tracking_res}
  % \vspace{-3mm}
\end{figure}

\textbf{Results}: By locating the TCP first in front of the camera at a fixed depth, we got the tracking results illustrated in Fig. \ref{tracking_res}. We notice that the tracking error on the x-axis has a mean value of $0.8 \ cm$, while the mean error on the y-axis is less than $0.7 \ cm$. The mean error on the z-axis was around $1.1 \ cm$.

As the quality of the detection decreases with the increased distance between the marker and the camera \cite{rabbi2014analysis}, we repeated the same experiment without fixing the depth of the TCP robot and recorded the errors on each axis. %, but this time we made the TCP go further away from the camera and recorded the errors on each axis.

\begin{figure}[h!]
  \centering
   \includegraphics[width=0.5\textwidth]{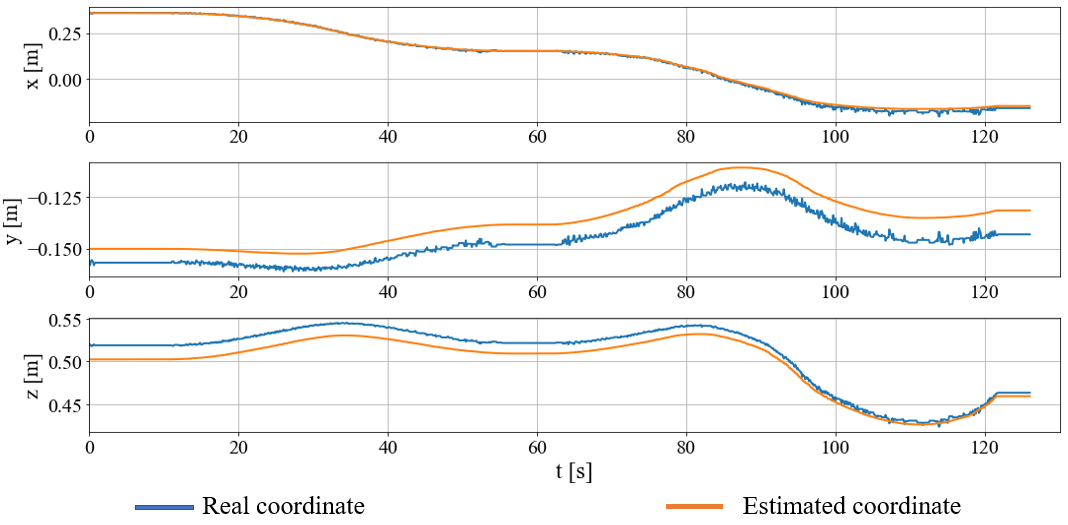}
  \qquad
  % \vspace*{-5mm}
  \caption{Experiment 1: Position error on each axis when increasing the distance between the marker and the camera along the camera z-axis.}
  \label{tracking_depth}
  \vspace{-3mm}
\end{figure}
We notice that the tracking error on the x-axis has a mean value around $0.5 \ cm$, while it was around $0.9 \ cm$ at the y-axis. The mean tracking error on the z-axis was around $1 \ cm$.
In order to assess the error in the position estimation of the marker radially from the robot TCP position, we observe the radial error for the two previous experiments which is defined as follows:
$$
e_{radial}=\sqrt{e_x^2+e_y^2+e_z^2},
$$
where $e_x$, $e_y$, and $e_z$ are the position error on the x-axis, y-axis, and z-axis respectively.
The results indicated that the radial error for both experiments ranged from $1.5 \ cm$ to $1.8 \ cm$. Based on this, we concluded that increasing the critical area diameter by $2 \ cm$ would ensure the safety of the operator's hand during the operation.  

The experiment shows that the tracking system is capable of dynamically tracking the operator's hand with an acceptable error.

\subsection{ Experiment 2}

The second experiment aims to evaluate the system's functionality in tracking the user's hand and adjusting the ArUcoGlide marker orientation for better detection. We also observed the behavior of the robotic arm driven by the collision avoidance algorithm in an imminent collision scenario. For this purpose, the robot is set to follow a path from point A to point B. A human operator wearing the device is moving his hand along and beside the robotic trajectory. The marker, while seen from the camera, will always give an update of the user's hand, and the orientation angles of the marker around both the y-axis and x-axis of the camera are being adjusted to guarantee better detection.

The critical distance of activation $d_{ACT}$ was set to 10 cm based on the tracking error (less than 2 cm) and the distance from the marker to the most distant point of the user's hand in a gripping posture. The avoidance threshold distance, $d_{AT}$, was chosen twice as the critical distance of activation plus a safety coefficient to account for potential abrupt movements of the user's hand. The haptic feedback was set to be activated in a range less than 30 cm, one motor is activated when the range is from $10 \ cm$ to $30 \ cm$, and both motors are activated when the range is less than 10 cm or when the marker is invisible to the tracking system. To further ensure safety, the system halts, and the robotic arm stops and moves into free mode in two situations. 1) when the marker is no longer detected and 2) when the operator's hand is closer than the critical distance of activation $d_{ACT}$ ($10 \ cm$) to the robotic arm. 

\begin{figure}[h!]
  \centering
  \includegraphics[width=0.45\textwidth]{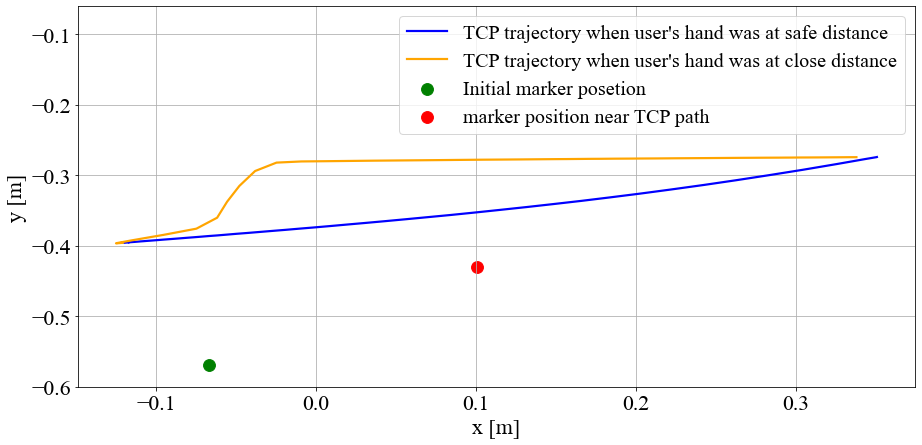}
  \qquad
  \vspace*{-5mm}
  \caption{Experiment 2: the TCP normal path when the user's hand was at a safe distance (blue line), and the TCP path curvature when the user's hand closely approaches the robot path (orange line)}
  \label{fig:dmin1}
\end{figure}

\textbf{Results}: Fig. \ref{fig:dmin1} shows the behavior of the path that the TCP has taken from point A to point B when the user's hand was at a safe distance (blue line). When the user approached his hand, the TCP path was curved to maintain a safe distance from the user's hand (orange line).  

The orientation of the marker can affect both the detection accuracy and the detection rate \cite{FiducialMarker1}. The device constantly adjusts the marker's orientation to ensure that the inclination angles around the y and x-axis of the camera frame remain close to the desired angles (40 degrees around the camera y-axis and 20 degrees around the x-axis). The value of the reference angle around the y-axis was chosen because it guarantees low error and high detection rates\cite{FiducialMarker1}. The reference for the x-axis was chosen at 20 degrees because it has a high detection rate on the one hand. On the other hand, this angle corresponds to an angular position of the upper motor at the center of its range of motion, hence more flexibility in adjusting the marker when the user moves his arm up and down without reaching a mechanical limit. Adjusting the orientation was achieved through a hysteresis controller, which guaranteed both angles stayed within a 10 degrees range centered around the desired values.
\begin{figure}[h!]
  \centering
  \includegraphics[width=0.45\textwidth]{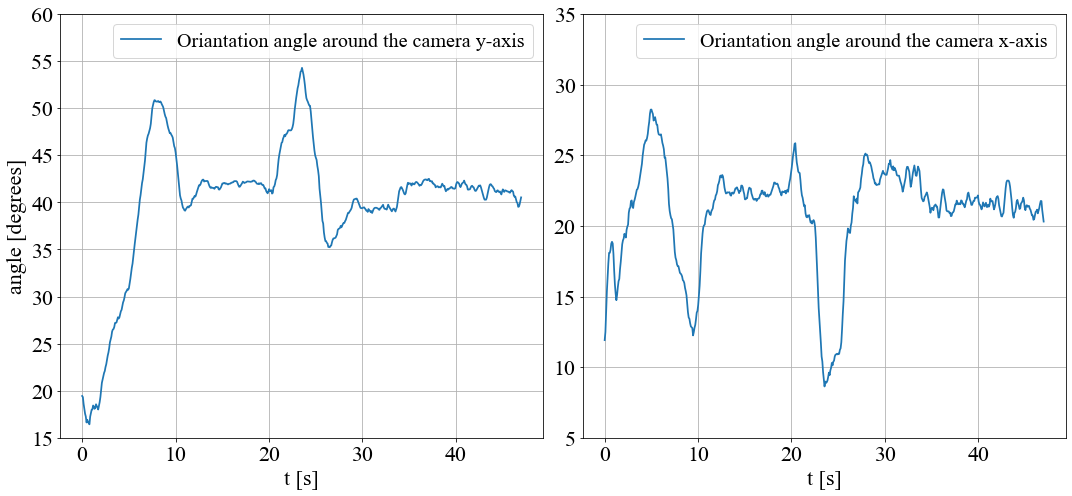}
  \qquad
  \vspace*{-5mm}
  \caption{Experiment 2: Marker orientation adjustment around the camera x-axis and y-axis.}
  \label{fig:orintation}
\end{figure}

Fig.\ref{fig:orintation} demonstrates the result of the marker orientation adjustment around the two mentioned axis during this experiment. we realize that both angles are stabilizing in a five-degree range around the desired angle. The picks in the two curves are related to the user's movement.

\section{Human-Robot Interaction Experiment in a Collaborative Task}

In this experiment, we aim to test the system in a real collaborative task to assess the integration of its parts in a real-world scenario.

\begin{figure}[ht]
  \centering
  \includegraphics[width=0.49\textwidth]{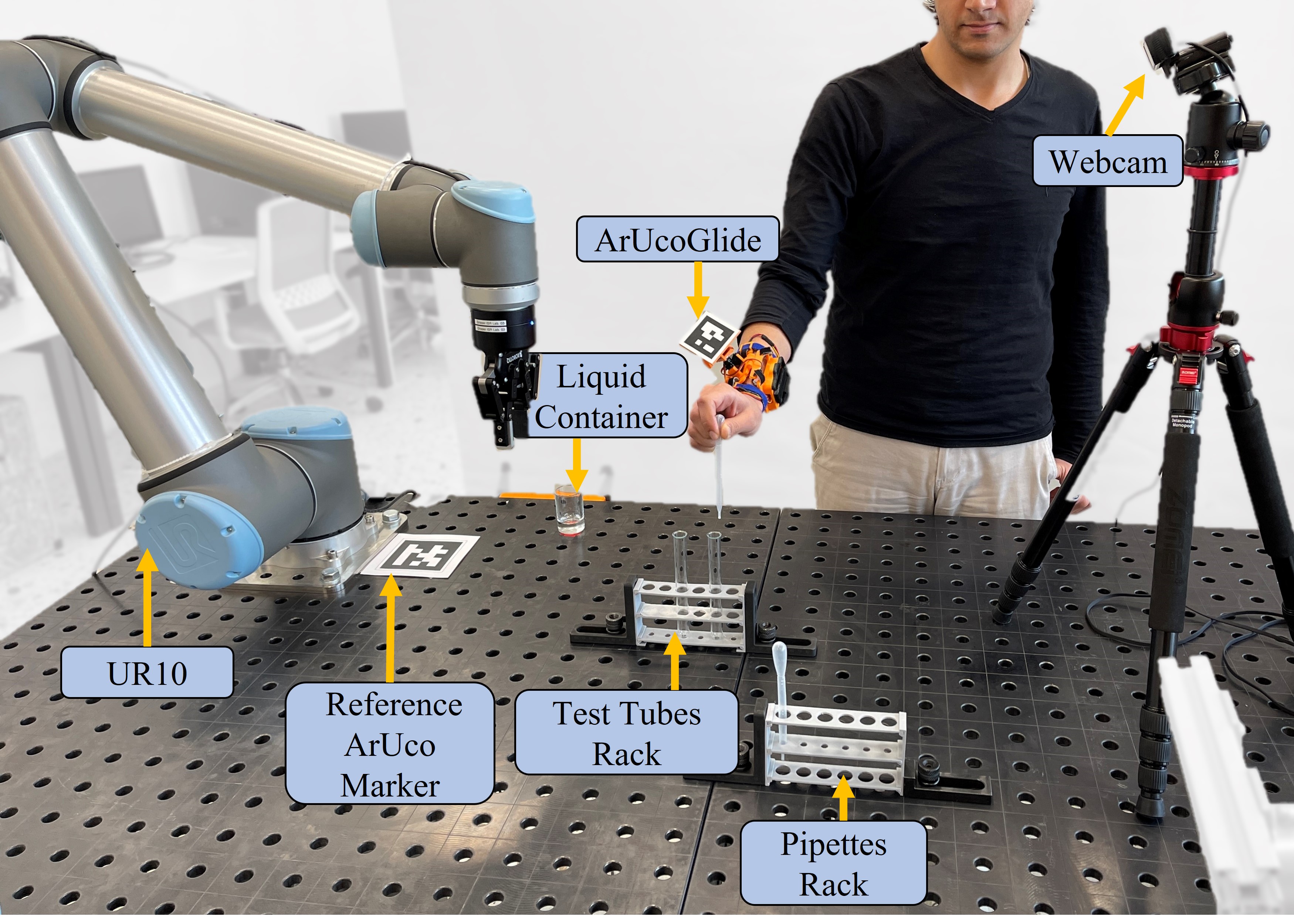}
  \qquad
  \vspace*{-5mm}
  \caption{Experimental setup. The robot takes empty pipette, fill it with liquid from the container and return it. The user counts liquid drops in a test tube. }
  % \vspace{-4mm}
  \label{fig:EnviromentExp3}
\end{figure}   

The objective of the experiment was to assess the performance of a human-robot collaboration system in a shared work environment reminiscent of typical activities found in warehouses or laboratories. The task for the robot (6 DoF collaborative robot UR10 with a 2-finger gripper from Robotiq) was retrieving an empty pipette from a pipette rack, moving towards a liquid-full container to fill the pipette with a certain amount of liquid, and returning it to its original position in the pipette rack. Meanwhile, the user was responsible for taking the filled pipette and counting the number of liquid drops it contained in a test tube, then emptying the test tube into a container outside the FOV of the camera. This scenario emulates a real experiment that can be conducted in a laboratory where the robot assists in delivering chemical liquids, either to speed up the process or to protect the user from exposure to toxic substances. We set up the place where the user does his task while wearing ArucoGlide device on his right forearm in the middle of the working space to intercept the robot's trajectory for better evaluation of the integration of the system.

During the experiment, four trials were conducted. In the first trial, the robot filled the pipette without any interference from the user. In the second trial, the user wore the device with the marker fixed parallel to their forearm throughout the operation without adjusting the marker's orientation. In the third trial, the device adjusted the marker's position to maintain a fixed orientation with the camera. Finally, in the fourth trial, haptic stimuli were activated along with adjusting the marker's orientation.

The critical distance of activation, $d_{ACT}$, was set to 10 cm, while the avoidance threshold distance, $d_{AT}$, was set to 30 cm. The maximum speed of the robot was set to 0.2 m/s, and the data collection frequency was set at 10 Hz. The haptic display was programmed to activate when the distance between the TCP and the marker was less than 30 cm. Initially, one vibration motor would be activated. However, as the TCP approached the critical area or when the marker became invisible, both vibration motors would be triggered simultaneously to alert the user to either move their hand away from the robot or adjust their hand position until the marker became visible again.

\begin{figure}[h!]

  \centering
  %\vspace*{5cm}
  \includegraphics[width=0.47\textwidth]{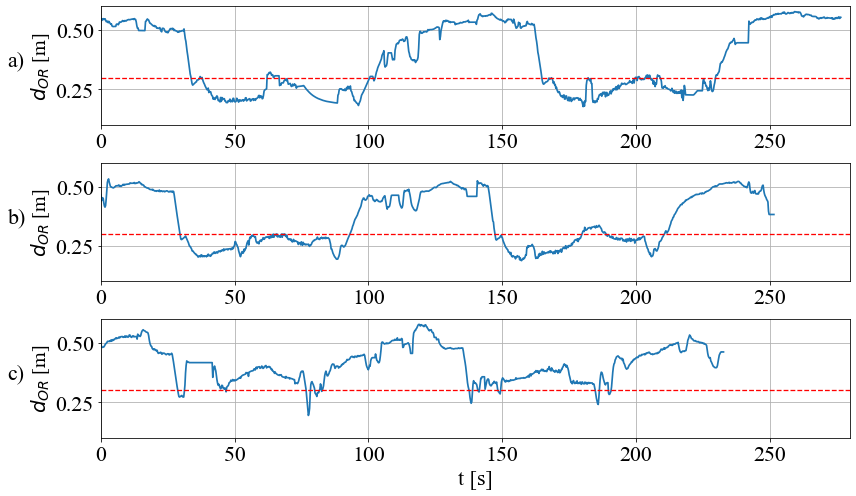}
  \qquad
  \vspace*{-0.8cm}
  \caption{Distance between the robot's TCP and the user's hand while ArUcoGlide is static (a), ArUcoGlide is adjusting the marker orientation (b), and ArUcoGlide is adjusting the marker orientation plus haptic feedback (c). The red line represents the avoidance threshold distance $d_{AT}$. }
  \label{fig:ex3}
  % \vspace{-3mm}
\end{figure}

\begin{figure}[h!]
  \centering
  \includegraphics[width=0.47\textwidth]{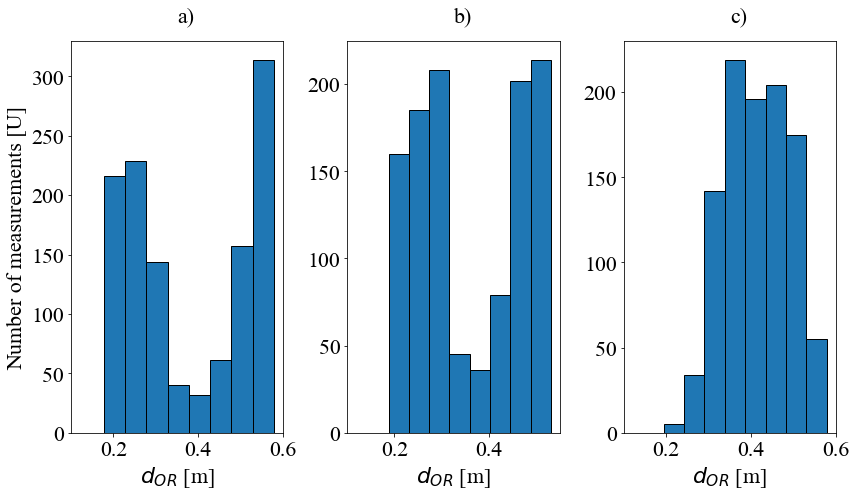}
  \qquad
  \vspace*{-5mm}
  \caption{Measurements of robot-obstacle distance inside of the range from 0 to 60 cm for the collaborative task while ArUcoGlide is static (a), ArUcoGlide is adjusting the marker orientation (b), and ArUcoGlide is adjusting the marker orientation plus haptic feedback (c).}
  \label{fig:dmin5}
  %\vspace{-3mm}
\end{figure}

\textbf{Results}: Fig. \ref{fig:ex3} shows the robot-obstacle distance $d_{RO}$ in the second (a), third (b), and fourth (c) mentioned trails. The minimum robot-obstacle distance in the second trial was around 17.5 cm, 18.5 cm for the third trial, and for the fourth trial, it was about 20 cm. The red line in Fig. \ref{fig:ex3} represents the avoidance threshold distance $d_{AT}$. Fig. \ref{fig:dmin5} shows the histogram of the robot-obstacle distance samples. We can observe that the number of measurements near the critical area has significantly reduced in the fourth trial (Fig. \ref{fig:dmin5} c)) compared to the other two trials. We realize in the fourth trial that the robot TCP has almost never been at a distance less than 20 cm from the user's hand. The haptic feedback was responsible for an increment in the robot-obstacle distance by $2.5 \ cm$. Also, the average robot-obstacle distance has increased by $5 \ cm$ due to haptic feedback.

\begin{figure}[h!]
  \centering
  \vspace*{0.2cm}
  \includegraphics[width=0.48\textwidth]{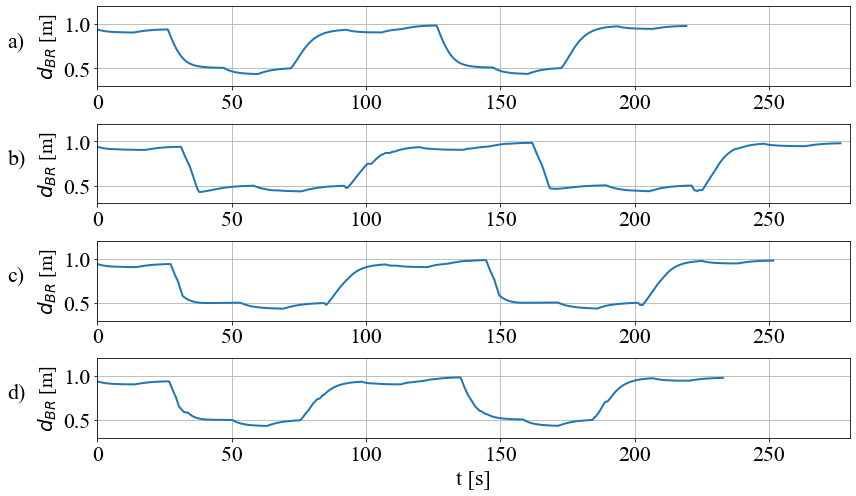}
  \qquad
  \vspace*{-5mm}
  \caption{TCP trajectories in the four cases: without obstacles (a), while ArUcoGlide is static (b), ArUcoGlide is adjusting the marker orientation (c), and ArUcoGlide is adjusting the marker orientation plus haptic feedback (d). vs. time. The trajectories 1, 2, 3, and 4 represent the TCP path to pick up each target (cube 1, cube 2, cylinder 1, and cylinder 2) and to transport them to the collecting box.}
  \label{fig:u}
  %\vspace{-3mm}
\end{figure}

Fig. \ref{fig:u} shows the distance from the TCP to the UR10 robot base. (Fig. \ref{fig:u} a)) shows the time the robot took to complete the task without any interference from the user. It took  219 s. We report the longest task time was for the second trial (fixed marker and without haptic feedback), which took around 276 s. The reason that the marker could get occluded is a result of the user's movement causing the robot to stop until the marker is visible again. In (Fig. \ref{fig:u} c)) we can see the benefit of adjusting the marker orientation as it reduces the moments of occlusion of the marker, resulting in a notable acceleration of the process to 251 s, marking a 9\%  improvement compared to the second trial. The employed time for the collaborative task with haptic feedback was 232 s, demonstrating an improvement of around 16\% from the third trial.

The recorded paths by the robot were 2.33 m for the first trial, 2.71 m for the second trial, 2.54 m for the third trial, and 2.36 m for the fourth trial. The collision paths (additional recorded paths by the robot to avoid obstacles) were  reduced significantly with haptic feedback to 3 cm from 38 cm in the second trail without haptic feedback (reduction of 92.1\%).

\section{Conclusion}

This research introduces an innovative framework for Human-Robot collaboration, which incorporates a wearable 2 DOF robot $ArUcoGlide$, a cost-effective and easy-to-install tracking system, and a collision avoidance controller.

The ArUcoGlide device aims to hold an ArUco marker and adjust the marker's position and orientation to maintain its visibility to the tracking system. It is also responsible for generating haptic stimuli to inform the user of hazardous situations.

Two experiments were conducted to assess the system functionality. The first, focused on measuring the accuracy of the tracking system, while the second experiment aimed to evaluate the behavior of the robot in an imminent collision scenario.

A third experiment was conducted to evaluate the integration of the whole system in a real-world collaborative task. We did three different trials: 1) without haptic feedback and with the marker fixed parallel to the user's hand, 2) with adjusting the marker orientation but without haptic feedback, and 3) with both adjusting the marker orientation and haptic feedback. Adjusting the marker orientation was responsible for speeding up the process by 9\%. While using the haptic feedback was responsible for speeding up the process by around 16\% and reducing the robot's path by  92.1\%.

In considering future work, we will prioritize downsizing the design of the ArUcoGlide to make it more comfortable for the user. It is also important to develop a mocap system that can capture the human body for safer interaction.

\addtolength{\textheight}{-12cm}   % This command serves to balance the column lengths

%%%%%%%%%%%%%%%%%%%%%%%%%%%%%%%%%%%%%%%%%%%%%%%%%%%%%%%%%%%%%%%%%%%%%%%%%%%%%%%%

%%%%%%%%%%%%%%%%%%%%%%%%%%%%%%%%%%%%%%%%%%%%%%%%%%%%%%%%%%%%%%%%%%%%%%%%%%%%%%%%

%%%%%%%%%%%%%%%%%%%%%%%%%%%%%%%%%%%%%%%%%%%%%%%%%%%%%%%%%%%%%%%%%%%%%%%%%%%%%%%%

%%%%%%%%%%%%%%%%%%%%%%%%%%%%%%%%%%%%%%%%%%%%%%%%%%%%%%%%%%%%%%%%%%%%%%%%%%%%%%%%

%\bibliographystyle{IEEEtran}
%\bibliography{bib}

\end{document}